\documentclass[acmsmall,screen,sigconf]{acmart}
\AtBeginDocument{%
  \providecommand\BibTeX{{%
    \normalfont B\kern-0.5em{\scshape i\kern-0.25em b}\kern-0.8em\TeX}}}

\newcommand{\RQA}{ How is intersectionality being applied to address bias in machine learning systems?}
\newcommand{\RQB}{How do prior efforts align and misalign with the theoretical foundations of intersectionality?}

\usepackage{txfonts}
\usepackage{wasysym}
\usepackage{booktabs}
\usepackage{colortbl}

\begin{document}

\title{A Preliminary Framework for  Intersectionality in ML Pipelines}

\author{Michelle Nashla Turcios}
\email{nashlaturcios@college.harvard.edu }
\orcid{0009-0006-2391-6646}
\affiliation{%
  \institution{Harvard University}
  \city{Boston}
  \country{USA}
}

\author{Alicia E. Boyd}
\email{alicia.boyd@yale.edu}
\orcid{0000-0002-2077-3823}
\affiliation{%
  \institution{Yale University}
  \city{New Haven}
  \country{USA}
}

\author{Angela D. R. Smith}
\email{adrsmith@utexas.edu}
\orcid{0000-0001-5546-5452}
\affiliation{%
  \institution{University of Texas at Austin}
  \city{Austin}
  \country{USA}
}

\author{Brittany Johnson}
\email{johnsonb@gmu.edu}
\orcid{0000-0002-0271-9647}
\affiliation{%
  \institution{George Mason University}
  \city{Fairfax}
  \country{USA}
}

\newcommand{\BJohnson}[1]{\textcolor{purple}{{\bfseries [[#1]]}}}
\newcommand{\ABoyd}[1]{\textcolor{blue}{{\bfseries [[#1]]}}}
\newcommand{\NTurcios}[1]{\textcolor{red}{{\bfseries [[#1]]}}}



\begin{abstract}

Machine learning (ML) has become a go-to solution for improving how we use, experience, and interact with technology (and the world around us).
Unfortunately, studies have repeatedly shown that machine learning technologies may not provide adequate support for societal identities and experiences. Intersectionality is a sociological framework that provides a mechanism for explicitly considering complex social identities, focusing on social justice and power. 
While the framework of intersectionality can support the development of technologies that acknowledge and support all members of society, it has been adopted and adapted in ways that are not always true to its foundations, thereby weakening its potential for impact.
To support the appropriate adoption and use of intersectionality for more equitable technological outcomes, we amplify the foundational intersectionality scholarship--Crenshaw, Combahee, and Collins (three C's), to create a socially relevant preliminary framework in developing machine-learning solutions. We use this framework to evaluate and report on the (mis)alignments of intersectionality application in machine learning literature. 
\end{abstract}




\maketitle

\section{Introduction}

In recent years, we have seen rapid growth in the popularity and use of machine learning technologies. 
From education to healthcare to the results we get from a web search, machine learning has become a go-to solution for an array of societal problems across a variety of domains and contexts~\cite{korkmaz2019review, shailaja2018machine, whittlestone2021societal}.
In the rapid adoption of machine learning solutions, we have seen promising outcomes regarding the ability for machine learning to improve the way we engage and interact with the world around us.
Most notable are advances in how we search for and retrieve information as existing search engines, like Google and Bing, have evolved and technologies like ChatGPT have emerged  that use deep machine learning to engage with end users~\cite{marr2019artificial, sun2023chatgpt, haleem2022era}.


The optimistic adoption of machine learning is not surprising given the potential to augment and even improve human efficiency~\cite{wan2019does, schwarzer2023bigger}.
Nonetheless, this widespread adoption and acceptance has not escaped criticism from practitioners and researchers who note the detrimental impacts of the biases in these systems~\cite{fuchs2018dangers, hildebrandt2021issue, angwin2022machine}.
Over the years, we have seen numerous examples of machine learning technologies disproportionately impacting historically marginalized communities for reasons such as lack of representation or explicit consideration of the potential for negative societal impact~\cite{ledford2019millions,obermeyer2019dissecting,hill2022wrongfully, chan2023harms}.
In light of this, scholars have commenced employing intersectionality as a theory-grounded methodology with the goal of informing their machine learning endeavors and thereby mitigating biases in outcomes. Broadly, the adoption of intersectionality has been used to address the existence of social identities to acknowledge their presence and work towards mitigating the harm of machine learning technologies. Intersectionality posits that understanding the experiences of individuals who identify with multiple social positions demands a departure from a one-dimensional lens, given the multidimensional nature of their positions and their location at the intersection~\cite{crenshaw1989demarginalizing, crenshaw1990mapping, collins2016intersectionality, collins2019intersectionality,collins_intersectionality_2015, collective1983combahee}. However, there are nuanced aspects of intersectionality that are not adopted holistically and often overlooked, such as interpretations of power and social justice~\cite{collins2019intersectionality,purdie2008intersectional,may2015pursuing, moradi2017using, del2021powering,bowleg2021master}. As a consequence, intersectionality and its potential for impact are weakened.  

To support researchers and practitioners interested in aptly adopting intersectionality to support the development of more equitable machine learning technologies, this paper provides a preliminary analysis of three existing efforts. 
Derived from foundational intersectionality scholarship~\cite{collective1983combahee, crenshaw1989demarginalizing,crenshaw1990mapping,collins_intersectionality_2015, collins2016intersectionality,collins2019intersectionality,crenshaw2010conversation}, we curated a preliminary set of guidelines to drive our analysis (further detailed in Section \ref{guidelines}). We then identified a small subset of case studies to evaluate. Our analysis provides an in-depth perspective of relevant considerations when applying intersectionality.
Moreover, we focused our analysis on three case studies that commonly cited three prominent intersectional scholars: 
the Combahee River Collective who integrated analysis into practice and posited that the major systems of oppression are interlocking~\cite{collective1983combahee}; legal scholar Kimberlé Crenshaw who introduced the term intersectionality when discussing the failure of U.S. law capturing Black women's experiences~\cite{crenshaw1989demarginalizing,crenshaw1990mapping}; 
and Sociology scholar Patricia Hill Collins who articulated the existence of multiple forms of oppression shifting the thinking of others beyond race, gender, and class to include sexuality and nation~\cite{collins_intersectionality_2015,collins2019intersectionality}.\footnote{As expressed by the AP style guide, throughout this paper, we capitalize ``Black.'' https://apnews.com/article/entertainment-cultures-race-and-ethnicity-us-news-ap-top- news-7e36c00c5af0436abc09e051261fff1f}
We found that, in all three case studies, while certain aspects of their work align well with foundation three C's (\textbf{C}ombahee River Collective, Kimberlé \textbf{C}renshaw, Patricia Hill \textbf{C}ollins), there exists gaps and misalignments in how their efforts discuss, interpret, and operationalize intersectionality.
More specifically, we found a number of methodological choices that diverged from the core tenets of intersectionality, which is rooted in Black feminism and social justice. 
Based on our analysis, we offer recommendations for future efforts in machine learning to ensure the contextually-aware and appropriate application of intersectionality.




\section{Background}


\subsection{Intersectionality's Foundations}

Before the term intersectionality was coined in the 1980s by legal scholar Kimberlé Crenshaw~\cite{crenshaw1989demarginalizing} as a way of understanding the experiences of Black women, earlier scholars of color, such as Harriet Jacobs, Audre Lord, Cherrie Moraga, Gloria Anzaldua, and Joseph Beam, shared writings on their intersectional experiences. Intersectionality is not a new concept but rather a way to acknowledge a longstanding phenomenon of people with multiple intersecting identities being confronted simultaneously through racism, imperialism, sexism, ability, nationalism, heterosexism, and colonialism~\cite{collins2016intersectionality, bowleg2016invited}. 
Over the years, the term has made its way through diverse academic disciplines, where discussions have emerged on how to use intersectionality as a theoretical framework~\cite{moradi2020mapping,bilge2013intersectionality}.
Hancock~\cite{hancock2013empirical}, argues that intersectionality does not fit a generalizable, validatable, explainable paradigm; instead, it is a critical theory. 
Critical theories aim to interrogate, expose, and explicitly challenge the status quo. Most importantly, critical theories aim to transform inequitable systems, not just ameliorate them~\cite{collins2019intersectionality, prilleltensky2009community,smith2018indigenous}. 
Therefore, intersectionality brings, as Collins describes it a `broad-based knowledge project'~\cite[pg.~3]{collins_intersectionality_2015}, which illuminates the critical praxis of social justice work. 

While the foundations of intersectionality focus on the intersection of race and gender (specifically for Black women), the critical theory of intersectionality encompasses many social positions beyond race and gender, including sexuality, age, disability, and immigration status~\cite{collins2019intersectionality, collective1983combahee,crenshaw1990mapping,carbado2013colorblind}. Disregarding of intersectional identities yields further marginalization within the realms of multiple oppressive systems (\textit{e.g.,} racism and sexism), as discussions will concentrate on addressing either one or the other, neither of which may genuinely represent their experiences.

\subsection{Intersectionality in Practice}

When considering the use of intersectionality as a critical theory, it is essential to view intersectionality as an analytical strategy. 
As highlighted by Patricia Collins, definitions of intersectionality should form the foundation of ``starting points for investigations rather than end points of analysis''~\cite[p.3]{collins_intersectionality_2015}. As an analytical strategy, prior work has primarily centered on intersectionality as a theory or a methodology in its practical application, or ``praxis.''

Despite the potential for intersectionality as a theory or methodology, prior efforts suggest methodological challenges when applying intersectionality in practice. This ranges from how to handle the complexity of identities concerning structural oppression~\cite{mccall2005complexity} and the improper use of mapping multiplication to social identities to understand complexity~\cite{hancock2007multiplication,bowleg2008black} to intersectionality transference from qualitative to quantitative practice~\cite{bowleg2013once, bowleg2008black}. 
In its growing popularity, we have seen the concept of intersectionality spread from its origins and morph into an unrecognizable theory that solely focuses (sometimes incorrectly) on complex social identities and neglects the considerations of power and social justice~\cite{collins_intersectionality_2015,bauer2014harnessing}.

Interpretations and applications of intersectionality have noticeably changed over the years, both in theory and practice, to the extent that many scholars believe it should even be considered a study of its own~\cite{bilge2013intersectionality}. 
Variations in terminology, definition, and methodology exist among scholarly discussions regarding intersectionality and its applications, creating a broad and sometimes disjointed community surrounding the theory~\cite{salem2018intersectionality}. 
Scholars caution against the excessive use of shorthand terms that invoke the complexity of intersecting power hierarchies, such as the phrase ``race, class, and gender,'' as frequent usage could diminish these terms to mere slogans. 
While ``intersecting system of power'' and ``intersecting systems of oppression'' tend to be used interchangeably, we should strive not to view them simply as descriptive concepts, as power and oppression and their impacts on individuals have far from a straightforward dynamic. While we recognize the dual purposes intersectionality can serve and the various interpretations that have emerged, our work emphasizes the three C's (Combahee, Crenshaw, \& Collins) definition of intersectionality as a methodology in quantitative studies, primarily due to its promising applications within machine learning.

\subsection{Intersectionality \& Machine Learning}

The rapid and widespread adoption of machine learning has brought about new relationships between the social and computational science research communities. 
More and more, we are seeing computing researchers look to insights from social and behavioral science researchers to better understand
and support more diverse users~\cite{kukafka2003grounding, pfleeger2012leveraging, hess2020sociotechnical, birhane2020towards, birhane2021algorithmic}.
One of the more recent examples in the context of machine learning is the integration of intersectionality into praxis~\cite{bowleg2016invited, Boyd2021QUINTA, szlavi2023intersectionality}.
This increasing attention to intersectionality (a social science concept) in machine learning points to a potentially mutually beneficial relationship between the two.


While the potential for synergy is apparent, machine learning has a long history of assuming binary attributes along a single axis.
Intersectionality, on the other hand, transcends binary considerations. Machine learning pipelines have generally failed to explicitly consider intersecting identities, despite tools like IBM's AI Fairness 360~\footnote{\url{https://ai-fairness-360.org/}} that provide mechanisms for exploring and mitigating bias in machine learning models. Traditionally, these pipelines focus on the existence or absence of individual attributes, such as being Black or white, with little attention paid to intersectional identities.



In the social science community, critical theories such as Black feminist theory, critical race theory, queer theory, and postcolonial theory offer explanations on why our technological systems consistently perpetuate societal inequalities~\cite{lugones2010toward, walter2013indigenous, moradi2017using, ogbonnaya2020critical}. 
Thus, many scholars have redirected their focus away from algorithmic fairness solely based on binary categories and towards broader considerations of equity and justice. 
Accordingly, the limitations of machine learning algorithms have been shown to disproportionately affect those who suffer from societal inequities, injustices, and exclusion from the dataset(s) being used~\cite{hill2022wrongfully,angwin2022machine,obermeyer2019dissecting}.
This issue extends beyond overall predictive accuracy and is known as ``predictive parity,'' or how the burden of predictive inaccuracy is distributed~\cite{dieterich2016compas}.

Given the potential for social science frameworks like intersectionality to have a positive impact on how we approach developing machine learning technologies, efforts like ours are important for supporting the proper adoption and use of social science frameworks like intersectionality.
In the next section, we describe the critical analysis we conducted of existing efforts that used intersectionality as a quantitative methodological approach. 

\section{Methodology}
\label{sec:methods}
The goal of our research is to amplify the considerations necessary for applying intersectionality in machine learning processes and provide insights into appropriate application based on existing efforts. Our efforts is driven by the following research questions:


\begin{description}
   \item[RQ1]\RQA
   \item[RQ2]\RQB 
\end{description}   





\subsection{Case Study Selection}

Our efforts began with a literature review on the global discourse surrounding the application of intersectionality in machine learning, reviewing 23 papers.
We found a dichotomy in the existing literature: certain works overtly and deliberately acknowledged Kimberlé Crenshaw in the context of discussing intersectionality; while others, though not explicitly citing Crenshaw, incorporated general nods to the concept of  intersectionality or acknowledged the theory’s consideration of multi-dimensionality without explicit reference.
Given the goal of our work to examine the sociological framework of intersectionality as defined by Crenshaw, we focused our critical analysis on the prior works that explicitly cited Crenshaw when defining intersectionality in their work.

Among the papers we collected, we found that there were three case studies in particular that represented the existing efforts at applying intersectionality in machine learning. We describe each of the prior works analyzed below.\\


\noindent\textbf{Case Study 1 (CS1)}\\
\noindent\textit{Towards Intersectionality in Machine Learning: Including More Identities, Handling Underrepresentation, and Performing Evaluation} ~\cite{wang2022towards}\\


CS1 investigated integrating intersectionality across various stages of the machine learning pipeline to understand the algorithmic effects of discrimination against demographic subgroups.
Using U.S. Census datasets, they focused on three stages: (1) determining which demographic attributes to include as dataset labels, (2) handling the progressively smaller size of subgroups during model training, and (3) moving beyond existing evaluation metrics for benchmarking model fairness across more subgroups.
The authors proposed the following considerations to be made at each of these stages: (1) comprehensive testing of granular identities and reallocating racial identities for specific groups (e.g., labeling Asian Pacific Islanders and advocating for a unique clustering technique for the heterogeneous ``other'' racial group) , (2) leveraging dataset structure to discern predictive patterns about a social group from others, and (3) correlating subgroup rankings based on positive label base rate and the rankings of true positives model predictions. Furthermore, the study suggested leveraging dataset structures to extract insights into underrepresented groups, such as Black women, by examining shared social identities with other groups, such as white women or Black men. 
Finally, their research proposes a novel fairness metric for evaluation that prioritizes equity (factors differences among people) over equality (which focuses on equal division, despite possible differences in need). 
This involves measuring the correlation between base rate rankings and True Positive Rates after applying a fairness algorithm, using Kendall's Tau. They aimed to unveil the perpetuation of social hierarchies within a given dataset.\\

\noindent\textbf{Case Study 2 (CS2)}\\
\noindent\textit{Critical Tools for Machine Learning: Working with Intersectional Critical Concepts in Machine Learning Systems Design}~\cite{klumbyte2022critical}\\


In CS2, the authors investigated the integration of intersectional critical theoretical concepts from social sciences and humanities into the design of machine learning systems. 
This exploration took the form of workshops that utilized intersectional feminist methodologies to craft interdisciplinary interventions for machine learning systems design, striving towards inclusivity, accountability, and contextual relevance.
The theoretical concepts of \textit{situating/situated knowledges}, \textit{figuration}, \textit{diffraction}, and \textit{critical fabulation/speculation} serve as the foundation for these concept-led design workshops. 
The paper outlined the design framework employed during the workshops, shedding light on the tensions and possibilities inherent in interdisciplinary machine learning systems design when the overarching goal is fostering more inclusive, contextualized, and accountable systems.\\

\noindent\textbf{Case Study 3 (CS3)}\\
\noindent\textit{Leveraging the alignment between machine learning and intersectionality: Using word embeddings to measure intersectional experiences of the nineteenth century U.S. South} ~\cite{nelson2021leveraging}\\

CS3 explored the intersectional experiences of individuals in the 19th century, emphasizing the synergy between machine learning and intersectionality by asserting an epistemological alignment. 
Using a dataset of nineteenth-century U.S. South narratives, the authors investigated word embeddings to better understand the experiences of Black men, Black women, white men, and white women within social institutions during this time period. The word embeddings were mapped to vectors representing American society's political, social, economic, and domestic spheres in the 19th century. Their findings revealed distinctions by race in the cultural and economic spheres within the corpus narratives. Gender became a distinguishing factor in the domestic sphere. The analysis indicates that Black men were afforded more discursive authority than white women within the examined texts.\\
\raggedbottom


\subsection{Analyzing Alignment with Theory} \label{guidelines}

To analyze the alignments of these case studies and their applications of intersectionality, we devised a set of guidelines, drawing inspiration from the scholarly contributions of Comhabee River Collectice, Kimberlé Crenshaw, Patricia Hill Collins, bell hooks, and Lisa Bowleg, Greta Bauer, Cherrie Moraga, Sirma Bilges, Gloria Anzaldua to name a few. 
More specifically, for each case study, we evaluated their methodological decisions based on the foundations and core tenets of intersectionality as described in these scholarly contributions.
We formulated insights from prior intersectionality scholarship into guidelines that can provide structure and facilitate our assessment of each case study's application of intersectionality.
Our guidelines, which aim to recognize the relational nature of intersecting social identities, dynamics of power, and entrenched inequalities, are described in detail below.\\




 \noindent\textbf{Guideline \#1 -- Intersectionality as a Relational Analysis}\\
 \textit{The machine learning project acknowledges and incorporates the relational nature of intersectionality within its data representation and feature engineering.}\\

We evaluated alignment with this guideline by determining if:
 \begin{enumerate}
     \item The project refrains from isolating social identities and instead recognize them as mutually constructive~\cite{collins_intersectionality_2015,collins2019intersectionality} 
     \item There exists modifications or creation of features that capture the interplay between different social dimensions~\cite{crenshaw1990mapping,moradi2017using}
     \item Intersectionality is integrated holistically into both data handling and analysis processes~\cite{bowleg2008black, hancock2007multiplication, Boyd2021QUINTA}
 \end{enumerate}

 \noindent\textbf{Guideline \#2 -- Social Formations of Complex Social Inequalities}\\
\textit{The machine learning project accounts for the social formations of complex inequalities resulting from intersecting systems of power in its data collection and preprocessing. The model appropriately handles imbalanced data and adequately represents diverse social experiences.}\\

We evaluated alignment with this guideline by determining if:
 \begin{enumerate}
     \item The approaches employed to tackle data imbalance actively acknowledge the presence of social inequalities~\cite{bauer2021intersectionality, mena2021classification,bowleg2013once}
     \item The methodologies handle imbalances with little to no  compromising of social identities~\cite{collins_intersectionality_2015, collins2019intersectionality}
     \item The project considers the emergence of social inequalities during data collection and processing~\cite{collins_intersectionality_2015, bauer2019methods,Boyd2021QUINTA, crenshaw1990mapping}
 \end{enumerate}

\noindent\textbf{Guideline \#3 -- Historical and Cultural Specificity}\\
 \textit{The machine learning project acknowledges the historical and cultural specificity of complex social inequalities during data gathering and model development. The training data is contextually appropriate and culturally sensitive to avoid perpetuating biases.}\\
 
We evaluated alignment with this guideline by determining if:
 \begin{enumerate}
     \item The work acknowledges the social justice facet of intersectionality~\cite{collins2019intersectionality, moradi2017using,hernandez1995re,bauer2021intersectionality}
     \item Researchers abstained from cursory mentions of intersectionality often limited to introductions, and instead, invoked it consistently throughout the entire paper ~\cite{collins_intersectionality_2015, moradi2020mapping,collins2019difference,may2015pursuing,hooks1981black}
     \item Researchers acknowledge the broader societal implications of their work~\cite{collins2019intersectionality,moradi2017using,alcoff1991problem,crenshaw2010conversation}
     \item Researchers contemplated cultural harm caused by reinforcing existing power structures~\cite{crenshaw1990mapping, collins2019intersectionality, collective1983combahee,alexander2012disappearing,cho_field_2013, bauer2014harnessing} 
 \end{enumerate}

\noindent\textbf{Guideline \#4 -- Feature Engineering and Statistical Methods}\\
\textit{The machine learning project's feature engineering and statistical methods account for intersectionality. The model considers the interactions between different social categories, avoids oversimplification, and incorporates intersectional variables in its analysis. }\\

We evaluated alignment with this guideline by determining if:
 \begin{enumerate}
     \item The project refrains from diluting social identities for computational convenience~\cite{hancock2007multiplication, bauer2019methods, bauer2014incorporating}
     \item Researchers avoid rigidly encapsulating individuals to fit into mathematical equations~\cite{hancock2007intersectionality,Boyd2021QUINTA, bauer2014incorporating, birhane2021algorithmic,saini2019superior, walter2013indigenous} 
     \item The work avoids diminishing the significance of people’s lived experiences~\cite{lugones2010toward,ortega2006being,anzaldua2004borderlands}
     \item The work considers the ramifications of mathematical methodologies in intersectionality, such as reduction of the complexity into simplicity~\cite{scheim2019intersectional, scheim2021advancing, bauer2019methods}    
 \end{enumerate}

\noindent\textbf{Guideline \#5 -- Ethical Considerations and Transparency}\\
\textit{The machine learning project demonstrates adherence to ethical guidelines and transparency regarding its use of intersectional analysis. The model's decisions are explainable, interpretable, and avoid reinforcing harmful stereotypes.}\\

We evaluated alignment with this guideline by determining if:
 \begin{enumerate}
     \item The project ensured replicability of methodologies employed~\cite{moradi2017using,bauer2019methods,guan2021investigation}
     \item Researchers describe if and to what extent replicating the work would avoid perpetuating harmful stereotypes~\cite{moradi2017using}
     \item The project established ethical parameters, such as researchers taking a reflexive position in the processing and  interpretation of their data and outcomes~\cite{collins2019intersectionality,bowleg2013once,boyd2021intersectionality,boyd2023reflexive}.
 \end{enumerate}




\section{Findings}

Our analysis focused on three case studies that explicitly and appropriately define intersectionality as defined by Combahee, Crenshaw and Collins~\cite{collins_intersectionality_2015,crenshaw1989demarginalizing,crenshaw1990mapping, collective1983combahee, collins2019intersectionality} and represent the few efforts being made at using intersectionality in machine learning research.
While all these case studies have commonalities in their efforts in integrating intersectionality into their machine learning pipeline, each project adopts unique modalities and serves disparate objectives.

\begin{table*}[ht]
\centering
\caption{Alignment with Intersectionality Across Case Studies}
\label{tab:alignment}
\setlength\tabcolsep{2pt}
\setlength\thickmuskip{0mu}
\setlength\medmuskip{0mu}
\begin{tabular}{l c c c}
\toprule
\textbf{Guideline} & \textbf{CS1}& \textbf{CS2} &\textbf{CS3} \\
\midrule
\rowcolor{gray!40}
(1) Intersectionality as a Relational Analysis & $\times$ & $\LEFTcircle$ & $\times$ \\
\hspace{3mm} (1.1) Recognizing mutually constructive nature of social identities & $\times$ & $\LEFTcircle$ & $\LEFTcircle$ \\
\rowcolor{gray!20}
\hspace{3mm} (1.2) Capture interplay between social dimensions  & $\LEFTcircle$ & $\times$ & $\LEFTcircle$ \\
\hspace{3mm} (1.3) Integration with data handling \& analysis  & $\LEFTcircle$ & $\times$ & $\times$\\
\rowcolor{gray!40}
(2) Formations of Complex Social Inequalities & $\times$ & $\times$ & $\times$  \\
\hspace{3mm} (2.1) Solutions for data imbalances acknowledge social inequalities & $\LEFTcircle$ & $\times$ & $\LEFTcircle$\\
\rowcolor{gray!20}
\hspace{3mm} (2.2) Little to no compromising of social identities & $\times$ & $\times$ & $\times$\\
\hspace{3mm} (2.3) Considerations for emerging social inequalities in data collection \& processing & $\times$ & $\LEFTcircle$ & $\LEFTcircle$\\
\rowcolor{gray!40}
(3) Historical \& Cultural Specificity & $\times$ & $\LEFTcircle$ & $\times$ \\
\hspace{3mm} (3.1) Acknowledges social justice facet of intersectionality & $\times$ & $\LEFTcircle$ & $\LEFTcircle$\\
\rowcolor{gray!20}
\hspace{3mm} (3.2) Intersectionality considerations integrated throughout reporting & $\times$ & $\times$ & $\times$\\
\hspace{3mm} (3.3) Acknowledges broader societal implications & $\times$ & $\medbullet$ & $\times$\\
\rowcolor{gray!20}
\hspace{3mm} (3.4) Contemplates cultural harm caused by reinforcing power structures & $\times$ & $\LEFTcircle$ & $\times$\\
\rowcolor{gray!40}
(4) Feature Engineering \& Statistical Methods & $\times$ & $\LEFTcircle$ &  $\times$ \\
\hspace{3mm} (4.1) Refrains from diluting social identities for computational convenience & $\times$ & -- & $\LEFTcircle$\\
\rowcolor{gray!20}
\hspace{3mm} (4.2) Avoid rigid encapsulation of individuals to fit mathematical equations & $\times$ & -- & $\LEFTcircle$\\
\hspace{3mm} (4.3) Avoids diminishing significance of lived experiences & $\LEFTcircle$ & $\LEFTcircle$ & $\times$\\
\rowcolor{gray!20}
\hspace{3mm} (4.4) Considers ramifications  & $\LEFTcircle$ & -- & $\LEFTcircle$\\
\rowcolor{gray!40}
(5) Ethical Considerations \& Transparency & $\times$ & $\times$ & $\times$ \\
\hspace{3mm} (5.1) Replicability of methodologies employed & $\LEFTcircle$ & $\LEFTcircle$ & $\LEFTcircle$\\
\rowcolor{gray!20}
\hspace{3mm} (5.2) Explicit discussion of replication's potential for perpetuating harmful stereotypes & $\times$ & $\LEFTcircle$ & $\times$\\
\hspace{3mm} (5.3) Established ethical parameters & $\LEFTcircle$ & $\times$ & $\times$\\
\bottomrule
\small{$\medbullet$ = aligned ; $\LEFTcircle$ = partially aligned; $\times$ = not aligned}
\end{tabular}
\end{table*}

\subsection{Alignment with Theory}
Using the aforementioned guidelines, we conducted a critical analysis of the methods applied in the selected case studies. Table ~\ref{tab:alignment} summarizes the case study's alignment with the specific guideline. As in the table, we denote alignment with a filled in circle ($\medbullet$), partial alignment with a half circle ($\LEFTcircle$), and non-alignment with an x ($\times$).
In this section, we outline the findings from our analysis. For each guideline, we outline the aspects we considered based on the enumerated lists in Section~\ref{sec:methods}.\\


\noindent\textbf{CS1: Intersectionality as a Relational Analysis} ($\times$)\hfill\\

\noindent\textit{Guideline 1.1 --} There was a misalignment in addressing the data imbalance. The researchers contradicted social identities being mutually constructive by arguing for extracting predictive patterns from similar groups to understand better (and inform other) underrepresented subgroups. For instance, the authors evaluated subgroup predictive disparities by training on select subgroups (Black female, white female, Black male) and testing on Black female. Therefore, their research suggested learning about Black females by observing patterns from Black males or white females, violating this guideline.\\


\noindent\textit{Guideline 1.2 --}
There is a partial alignment with this guideline due to the consideration of racial group granularities. The researchers delved into the granularity of constituent identities within the Asian Pacific Islander to substantiate their consideration of relevant identities in dataset labeling. However, when handling ``Other'' racial groups, the authors used redistributive mathematical techniques to assign the racial group to its closest neighbor in the feature space. This process oversimplified racial identities and combined more vulnerable communities with less vulnerable ones.\\


\noindent\textit{Guideline 1.3 --}
We observed partial alignment with the authors' attempt to study granular data of racialized groups and with how to apply intersectionality holistically. Yet, we caution against marginalizing vulnerable groups by merging them with dominant ones. The researchers' method oversimplified intersectional identities and lacked sensitivity to their complex interplay.\\

\noindent\textbf{CS1: Formations of Complex Social Inequalities} ($\times$)\\

\noindent\textit{Guideline 2.1 --}
We found a partial alignment with this guideline. The authors attempted to look at heterogeneity within a single axis, race. However, the authors posit that specific racial subgroups may exhibit sufficient homogeneity to warrant aggregation. \\

\noindent\textit{Guideline 2.2 --}
We found no alignment with this guideline due to the researchers reallocating individuals within specific racial groups to those that exhibited the closest proximity in the feature space, directly compromising their identities. There was a nuanced experiential difference and the unintentional reinforcement of oppression when combining underrepresented and dominant groups.\\

\noindent\textit{Guideline 2.3 --} 
Given the fusion of racial identities, there was no alignment here. The researchers neglect the ramifications of additional identity axes (\textit{i.e.,} non-able biracial individuals) and do not consider the nuanced aspects of racialized burdens.\\



\noindent\textbf{CS1: Historical and Cultural Specificity} ($\times$)\hfill\\

\noindent\textit{Guideline 3.1 -- }
We found misalignment for this guideline -- the researchers failed to demonstrate cultural sensitivity at various junctures within their project’s pipeline.\\

\noindent\textit{Guideline 3.2 -- }
We determined there was misalignment due to the researchers' approaches which were geared towards mitigating algorithmic harm within a theoretical space. However, there was no apparent connection between theoretical constructs and practical application. The research approach failed to acknowledge the social justice aspects of intersectionality.\\

\noindent\textit{Guideline 3.3 -- }
There was no alignment with this guideline. The authors noted a lack of concern about social identity aspects of intersectionality -- instead, they were narrowly focused on the quality of their unfairness metrics. This rendered the use of intersectionality inappropriate, given that intersectionality must acknowledge the broader societal implications and power dynamics. The researchers, for instance, prioritize the exploration of heterogeneity within groups along a singular axis, contending that unity emerges not solely from shared traits \textit{but} from a collective experience of adversity driving united change. This perspective implicitly acknowledges underlying inequalities within these groups. Notably, this notion holds in tangible instances, such as the heterogeneous nature of the Latino identity. Additionally, their examination of the heterogeneous “Other” group recognizes the real-world dynamics that this amalgamation encompasses, drawing adequate attention to varied forms of discrimination its constituents face. \\

\noindent\textit{Guideline 3.4 -- }
There was a misalignment with this guideline, because the researchers' quantitative approach did not encompass the nuanced, intersectional dimensions specific to complex social identities' experiences (\textit{i.e.,} Black female), ultimately reducing complex social identities to accommodate them within a mathematical framework or machine learning model for computational convenience. Furthermore, the researchers' emphasis on mathematical vs. societal structure bypassed insights from social sciences, hindering the recognition of historical complexities.\\

\noindent\textbf{CS1: Feature Engineering and Statistical Methods} ($\times$)\hfill\\

\noindent\textit{Guideline 4.1 -- }
We found misalignment with this guideline based on the reliance solely on shared characteristics (which risks losing granularity), and the exclusive focus on predictive differences which oversimplified the fairness analysis. This was observed when the researchers' proposed technique of optimizing a white female to Black male training sample for the highest AUC on a Black female validation set. Hence, this proposed technique obscured the nuances specific to particular intersectional identities, particularly Black females.\\

\noindent\textit{Guideline 4.2 -- }
We observed no alignment with this guideline. The researchers' strategies to integrate intersectionality into their machine learning project highlighted a fundamental deficiency: attempting to represent human beings and their social identities with mathematical constructs. Throughout the paper, the researchers underscored the significance of domain expertise and contextual information in shaping algorithmic and mathematical determinations. However, they paradoxically defied these assertions by forcing individuals into inflexible frameworks, often without substantial grounding in social science, yielding erroneous conclusions. \\

\noindent\textit{Guideline 4.3 -- }
We observed a partial alignment. The researchers recognized the limitation of extrapolating existing fairness metrics as subgroups increased beyond binary settings. This awareness reflected the acknowledgment that intersectionality surpasses mere binary categorizations. However, the researchers' research questions did not account for intersectional lived experiences.\\


\noindent\textit{Guideline 4.4 -- }
We found partial alignment with this guideline. The researchers initially acknowledged the misalignment of intersectionality with the fairness metrics and criticized the lack of applicability across social subgroups. However, the researchers' definition and approach of hierarchy and its relationship to social identities and power dynamics played into the complexity and avoided simplification resulting from the use of statistical measures.\\

For scenarios with numerous subgroups, the researchers propose a new metric assessing the correlation between base rate and TPR rankings post-fairness algorithm application. Employing statistical measures like Kendall’s Tau and p-values objectively examines intersectional hierarchies, aiding in identifying disproportional effects and hierarchy disruptions.  By acknowledging the need for improved intersectional metrics, the researchers contribute to the emerging field of intersectional statistics, enhancing overall bias assessment within their model.  \\

\noindent\textbf{CS1: Ethical Considerations and Transparency} ($\times$)\hfill\\

\noindent\textit{Guideline 5.1 -- }
We found partial alignment with this guideline as the researchers used graphical representations to illustrate findings extensively and substantiated corresponding claims. However, the replication of this work warrants caution to accept these findings as computational accuracy did not necessarily translate to real-life accuracy.\\


\noindent\textit{Guideline 5.2 -- }
We determined that this guideline was not met, as replicating this work would contribute to harmful stereotypes. Furthermore, the work's deficiency became particularly evident when the researchers generated knowledge that contradicted one of intersectionality's fundamental tenets: namely, the notion that social positions are mutually constitutive (Guideline 1.1).\\


\noindent\textit{Guideline 5.3 -- }
We found partial alignment with this guideline because the project’s decisions were presented with a purported level of transparency, accompanied by an explication of the underlying rationale. However, a critical examination of the researchers' position revealed that the logical underpinning of these decisions was flawed, potentially yielding adverse outcomes upon replication influencing the interpretation of their findings.\\





\noindent\textbf{CS2: Intersectionality as a Relational Analysis} ($\LEFTcircle$)\\

\noindent\textit{Guideline 1.1 -- }
We determined there was a partial alignment with this guideline due to the researchers grounding their work, citing Collins~\cite{collins1991black} discussion on 'situated knowledge' from specific standpoints. However, the relationality discussion needed to be included in the standpoint conversation, which Collins~\cite{collins2019intersectionality} discussed in great detail, recognizing social identities as mutually constructive. The shortfall lies in the absence of a discernible indicator of whether the term ``positionality'' employed by the researchers encompasses social identities—fundamental to intersectionality's scope.\\


\noindent\textit{Guideline 1.2 -- }
We found no alignment with this guideline; the researchers did not explicitly reference social categories or delineate the role of social power within the workshop. Therefore, an inadequately established linkage exists between intersectionality scholars' conceptualization of relationality and the researchers' ``situation'' concept definition.\\

\noindent\textit{Guideline 1.3 -- }
There was no alignment with this guideline due to the authors' abstraction, unclear explanation, and lack of application of intersectionality, therefore signifying the under-utilization of intersectionality's full potential, hindered the assessment of social inequalities within the data.\\ 


\noindent\textbf{CS2: Social Formations of Complex Social Inequalities} ($\times$)\\

\noindent\textit{Guideline 2.1 -- } 
There was no alignment with the guideline due to the workshop's limited utilization of intersectionality, reflecting a scenario wherein intersectionality was not fully harnessed due to misconceptions or knowledge gaps regarding its implementation. It remains to be seen whether the figuration methodology used comprehensively addressed this perspective. This would underscore the interdependence between users and technical systems and help counter the misconception that users and technical systems are distinctly separate. This is significant since this misconception impacts the acknowledgment of social inequalities in data.\\



\noindent\textit{Guideline 2.2 -- }
There is no alignment to this guideline due to the under-application of intersectionality, which is estranged from its original context. Therefore, the paper failed to establish a coherent connection between the methodology and the foundational principles of intersectionality, thus harming social identities and power dynamics.\\



\noindent\textit{Guideline 2.3 -- }
We observed partial alignment for this guideline. The authors' conceptualized the power hierarchies inherent in social formations, consequently giving rise to multifaceted inequalities via pedagogical means. However, there was a missing consideration for the impact of these systems on social identities and positional power, which have impacted data processing.\\


\noindent\textbf{CS2: Historical and Cultural Specificity} ($\LEFTcircle$)\\

\noindent\textit{Guideline 3.1 -- }
For this guideline, we found a partial alignment due to the authors' acknowledging the lineage of Black critical theoretical thought in intersectionality, particularly their mention of power hierarchies and absent perspectives. While the researchers mentioned the importance of historical underpinnings, there was no explicit application nor acknowledgment of the historical impacts on the social justice facet of intersectionality. \\


\noindent\textit{Guideline 3.2 -- }
There was no alignment with this guideline because the researchers did not abstain from cursory mentions of intersectionality, consistently limiting the theory to the introductions.\\

\noindent\textit{Guideline 3.3 -- }
We observed alignment with this guideline due to the researchers' acknowledgment of the broader societal implications of their work, \textit{e.g.,} considering how machine learning systems have potential harms and benefits. The societal implications were demonstrated by encouraging participants to contemplate user experiences beyond their own, fostering cultural sensitivity, and transcending the focus on individual needs. This approach manifested as a deliberate discourse integration to address the absence of perspectives, representing a more apt approach for evaluating equity than traditional fairness evaluation metrics.\\

\noindent\textit{Guideline 3.4 -- }
We found this guideline had partial alignment. The researchers actively engaged with historical and cultural specificity, encapsulating the project's dedication to contextual relevance and cultural sensitivity. However, the researchers failed to address the cultural harm caused by reinforcing existing power structures.\\

\noindent\textbf{CS2: Feature Engineering and Statistical Methods} ($\LEFTcircle$)\\

\noindent\textit{Guideline 4.1 -- }
This guideline was not applicable due to insufficient knowledge on whether the project refrained from diluting social identities from computational convenience. The paper centered on machine learning design and did not mention mathematical methods deployed or algorithms used.\\

\noindent\textit{Guideline 4.2 -- }
This guideline was not applicable due to insufficient knowledge to determine whether the researchers avoided rigidly encapsulating individuals to fit mathematical equations. As noted in the previous guideline (4.1), the paper centered on design and did not mention mathematical methods deployed or algorithms used.\\

\noindent\textit{Guideline 4.3 -- }
This work partially aligned with this guideline due to the researchers explicitly and intentionally not diminishing the significance of people's lived experiences. However, per Guideline 3.2 of this case study, there needed to be an explicit mention of intersectionality, which invokes the direct power constructs of people's lived experiences.\\



\noindent\textit{Guideline 4.4 --}
This guideline was not applicable due to insufficient knowledge on whether work considered the ramifications of mathematical methodologies in intersectionality, such as reducing complexity into simplicity. The paper centered on machine learning design and does not mention mathematical methods deployed or algorithms used.\\


\noindent\textbf{CS2: Ethical Considerations and Transparency} ($\times$)\\

\noindent\textit{Guideline 5.1 -- }
There was partial alignment with this guideline due to how the researchers structured their workshops and posed questions to the participants about the societal impacts of machine learning systems. However, replication would be difficult due to the disconnect between intersectionality and its foundations.\\


\noindent\textit{Guideline 5.2 -- }
We observed partial alignment to this guideline. The work established the foundational scholarship from Collins to adopt equitable design principles for machine learning systems. However, there was no robust implementation nor explanation of how intersectionality and power dynamics are deeply rooted in critical feminist, anti-racist, and postcolonial scholarship, which the authors list but fail to link to the harmful replication of stereotypes in their work.\\


\noindent\textit{Guideline 5.3 -- }
There was no alignment with this guideline. The researchers' did not mention default privilege from a non-U.S. context and a lack of reflexive mention of colonialism and imperialism, which intersectionality's reflexive practice would have illuminated if intersectionality had been actually implemented correctly, per Guideline 1.1.\\

\noindent\textbf{CS3: Intersectionality as a Relational Analysis} ($\times$)\\

\noindent\textit{Guideline 1.1 -- }
We observed partial alignment with this guideline, given the principle of relationality where the author establishes a foundation for the mutually constructive interrelation within social positioning. However, the results and analysis diverged from the reality of social and power constructions of minority groups from the 1800s.\\

\noindent\textit{Guideline 1.2 --}
There was partial alignment with this guideline through its use of qualitative data via word embeddings on par with inductive logic. It underscores the conformational analysis of social positions, highlighting their original and mutually constructive nature. However, there was a divergence between the results and analysis by imposing multi-dimensional machine learning techniques to map explicitly to intersectional identities without acknowledging the nuances of structural political powers of complex social identities.\\


\noindent\textit{Guideline 1.3 --} 
We observed no alignment as the author determined gender assignments by examining first names or by scrutinizing biographical details when ambiguity persisted. Further, it is noteworthy that the meaning of queer identity in the nineteenth century differed significantly. We must be cautious of inadvertently adopting a heteronormative view of gender, possibly neglecting individuals at the margin of this spectrum.\\

\noindent\textbf{CS3: Social Formations of Complex Social Inequalities} ($\times$)\\

\noindent\textit{Guideline 2.1 -- }
We found this guideline in partial alignment due to this project establishing a nuanced context by delineating four distinct institutional spheres: culture, economy, politics, and domestic. However, an explanation of how the author accounted for data imbalances within word embeddings and creation must be explained. Further, the author's biases within the word embedding models to analyze the different century periods and social groups can marginalize social identities.\\


\noindent\textit{Guideline 2.2 -- }
We found no alignment with this guideline. More specifically, the author referenced the utilization of machine learning’s multidimensional scaling capabilities to offer a more comprehensive perspective on intersectionality. Their assertion hinged on the capacity to incorporate multiple identity axes and represent various systemic facets. The author further aligned their approach with the contention that machine learning seamlessly fuses inductive logic with 11rich and complex representations of traditionally qualitative data.'' The author's presumed a seamless transference of machine learning logic onto intersectional logic. Unlike machine learning, intersectionality demands human intervention, as it is grounded in lived human experiences.\\


\noindent\textit{Guideline 2.3 -- }
We observed partial alignment based on the researcher's attempts to adopt a qualitative data-driven approach. However, the approach failed to account for the historical context appropriately and needed to improve in facilitating the contextualized positioning of word vectors. In a broader context, the incorporation of word embeddings might typically accentuate the pertinence of qualitative data; however, in this case, it failed without the appropriate historical context.  \\

\noindent\textbf{CS3: Historical and Cultural Specificity} ($\times$)\\

\noindent\textit{Guideline 3.1 -- }
We observed partial alignment with this guideline. The author traced the lineage of intersectionality (i.e., cited the foundational scholars Combahee, Crenshaw, and Collins) back to anti-slavery movements, thereby increasing the obligation to work alongside the framework with heightened responsibility. However, there was not more intentional intersectional application and acknowledgment of intersectional experiences of marginalized groups and insights based on intersectionality's citational lineages. Otherwise, intersectionality can be rendered performative.\\

\noindent\textit{Guideline 3.2 -- }
We found no alignment with this guideline. The study lacked a rigorous historical and cultural examination of social identities. This, coupled with the biased nature of social identity selection, undermined the project's ability to capture the genuine sociocultural inequalities of the era. The author's classification of the race variable relied on the inclusion of documents in the “North American Slave Narratives.” When clarity was lacking in these criteria, the author analyzed the documents themselves to discern if they pertained to or were authored by Black individuals. This approach raised a critical concern: \textit{Did it inadvertently employ a white colonial lens in the textual analysis?} This question is particularly pertinent considering that many slaves during that historical period lacked the literacy skills necessary for reading and writing. These points necessitate careful consideration when evaluating the application of intersectionality as a relational analysis. \\

\noindent\textit{Guideline 3.3 -- }
We found no alignment with this guideline. The author advised interpreting her findings as representative of a specific facet of nineteenth-century U.S. discourse, primarily focused on the daily experiences of individuals during that era and leaning toward abolitionist sentiments. However, her findings did not encompass the heterogeneous nature of the selected documents, spanning various dates throughout the nineteenth century. Further, the diverse temporal range of the corpus introduced conflicting ideologies, linguistic shifts, and societal changes, which could compromise the reliability and interpretability of the results by failing to acknowledge the disconnected societal implications of her work in the current day.\\

\noindent\textit{Guideline 3.4 -- }
We observed no alignment with this guideline due to the author's intentionality in selecting documents. The author articulated: “To re-construct these diverse experiences of slavery and reconstruction, historians have scoured the archives, reinterpreting preserved documents and republishing documents that were almost lost to history.” Nevertheless, a critical examination emerged when one questions the author’s choice of this corpus and the rationale behind it. The author maintained: “These documents were purposefully selected by the collection curators to convey the everyday experience of those living in and around the U.S. South whose voices are often not heard, or not heard as loudly.” This intentionality in selecting documents, invites scrutiny. Historians, sociologists, and others within the academic realm are fundamentally tasked with preserving and representing the voices of the past, not with creating or imposing voices. The author’s approach risked inadvertently silencing these voices by speaking on their behalf. This phenomenon can lead to an incomplete representation of these individuals’ authentic lived experiences, potentially robbing these narratives of their cultural significance.\\


\noindent\textbf{CS3: Feature Engineering and Statistical Methods} ($\LEFTcircle$)\\

\noindent\textit{Guideline 4.1 -- }
We noted partial alignment with this guideline because the author acknowledges a disconnect between traditional inferential statistics within an intersectional context. The author attempted to address the quantitative limitations within intersectionality; she used word embeddings to circumvent the limitations of mathematical constraints. However, characterizing machine learning as ``radically inductive'' and ``foundationally compatible'' with intersectionality is overly reductionist towards both concepts. Intersectionality, as an epistemological paradigm, accords precedence to qualitative and experiential knowledge, whereas machine learning predominantly relies on quantitative data.\\

\noindent\textit{Guideline 4.2 -- }
We found partial alignment for this guideline because the researcher highlighted a distinctive strategic decision: using word embeddings to represent qualitative data. This choice emphasized the prioritization of qualitative data. However, the researcher’s word embedding methodology has limitations, primarily rooted in data collection methodologies and unaddressed ethical considerations.\\


\noindent\textit{Guideline 4.3 -- }
We observed misalignment with this guideline due to the ethical tensions of not consulting with the corpus work writers about their intended meaning. The author bore a heightened responsibility to refrain from narrating a story that fails to authentically reflect the intersectional experiences of the real people behind the corpus content. This diminishes the significance of their lived experiences.\\

\noindent\textit{Guideline 4.4 -- }
We observed partial alignment with this guideline because the author's earlier distinction that intersectionality cannot be studied from a purely quantitative stance. However, there are underlying mathematical components behind the creation of word embedding models that the author does not consider, which further reduces the intersectional complexity of the data. \\


\noindent\textbf{CS3: Ethical Considerations and Transparency} ($\LEFTcircle$)\\

\noindent\textit{Guideline 5.1 -- }
We observed partial alignment due to the opportunity to replicate the author's work to garner findings. Yet, the replication of this project without refinements poses the risk of generating knowledge susceptible to fostering inaccurate inferences regarding the interplay between social identities and power dynamics among social groups (\textit{i.e.,}  Black men and white women).\\


\noindent\textit{Guideline 5.2 -- }
We found no alignment with the guideline. Specifically, the default privileges of a researcher were inherently influential in shaping the interpretation of discursive meaning embedded within word vectors. Therefore, a prudent approach is warranted to discourage the utilization of writings from marginalized groups during periods of oppression towards generating knowledge that may lack a comprehensive representation of their experiences.\\


\noindent\textit{Guideline 5.3 -- } 
We observed a misalignment with this guideline because of the author's reflexive positioning. The author does not apply reflexive analysis to accurately represent word embedding methodology consistently. The author's reliance on the word embedding model output for validating findings needs more reflexivity based on the structural underpinnings of the work generated by machine learning algorithms. \\


\section{Integrating Intersectionality}
Our study provides a glimpse into the ways in which we can apply intersectionality in the context of designing and developing machine learning technologies and systems. 
While there is interest in adopting intersectionality as a methodology in machine learning processes, prior work along with our analysis of existing efforts amplify the concerns regarding its use in practice~\cite{bauer2021artificial, bauer2021intersectionality,bauer2022latent,bauer2019methods,mahendran2022describing,mahendran2022quantitative}. In this section, we discuss the ways in which we can  apply intersectionality in technical or quantitative processes without misrepresenting its intent or diminishing its potential for positive impact.

\subsection{Agenda Alignment}
While the complexity of social identities is central to intersectionality as a critical theory, it is not the sole tenet. 
Intersectionality promotes a social justice agenda that is critical to proper application and interpretation, especially (as mentioned above) when citing foundational scholars such as Cohambee, Crenshaw, Collins (or three C's)~\cite{davis2008intersectionality, moradi2017using,salem2018intersectionality,bowleg2021master}.
When adopting intersectionality to address inadequacies in machine learning processes and outcomes, 
it is important to realize and align with the social justice agenda that is fundamental to its definition.

It may feel that simply stating that the work being done is not directly exploring societal impacts or implications would be enough to avoid this criticism.
However, even with such a declaration, the potential for societal impacts are implied and, if adopted without proper consideration, inevitable.
Our work suggests there may exist efforts, like CS2, that at least acknowledge and articulate their goal of fostering less biased machine learning systems. However, we also uncovered  efforts like CS1 and CS3 that attempt to explicitly declare lack of intent regarding societal implications or propagate historically inaccurate assertions that fail to consider real life experiences and potential impact.


Our observations prompt an important query: \textit{why reference intersectionality, a critical theory centered on social justice, without the explicit consideration for societal implications or impact?}
Given the growing interest in and ubiquity of machine learning technologies, we posit that any effort in research or practice that focuses on advances in machine learning \textit{must} consider the implications of their contribution for \textit{all} potential stakeholders. 
This is especially the case when the efforts are focused on populations that we know disproportionately suffer from both societal and technological inequities and that the potential for impact, positive or negative, is so real~\cite{hill2022wrongfully, angwin2022machine, obermeyer2019dissecting}.
By intentionally aligning with intersectionality's societal considerations, researchers have the opportunity to provide insights and solutions that are not only theoretically sound, but also applicable in practical contexts.

%

\subsection{Power \& Positionality}
As researchers, we often find ourselves having to interpret and represent perspectives and experiences that are not our own.
With that, it is our responsibility to ensure that we are making all due considerations in our interpretations.
One of the ways researchers add perspective and context to their interpretations of their efforts is by providing a positionality statement~\cite{bourke2014positionality, hampton2021positionality, martin2022positionality}. 
Positionality statements provide researchers with an opportunity to be reflexive regarding their own position and its role in their ability (or lackthereof) to adequately and accurately interpret the insights from their findings.
In our sample, CS1 provided a positionality statement that acknowledged their position and how they may be in opposition to the positionality of the groups they are investigating in their research.

While positionality statements help contextualize the interpretations reported, this is only one part of a larger process to ensure the propagation of insights that are accurate and more importantly not harmful~\cite{savolainen2023positionality}.
To be truly reflexive requires due diligence in the process of analyzing data and interpreting findings; this includes, but is not limited to, reflecting on how the ways in which the methods or interpretations consider or address social identities, privilege, and power dynamics~\cite{martin2022positionality, jones2010putting,collins2019intersectionality,boyd2023reflexive}.
This is underscored by the power researchers have in shaping the discourse around innovation. Given a researcher's professional and social position, their assertions may be deemed credible (and reused throughout the community) at the cost of marginalizing the experiences of those they aim to support.
Consider CS3, the researcher interpreted their findings through a position of privilege and failed to acknowledge how said privilege might affect their interpretation. Furthermore, they used their position to unintentionally promote a narrative that contradicts lived experiences of a marginalized sub-population (and can be cited by others as having scientific backing). The authors of this paper are all women of color and hold multiple marginalized identities. The first three authors are deeply involved in intersectionality scholarship. As such, our lived experiences shape our understanding of the literature and the considerations surrounding  intersectionality.

When wielding intersectionality in machine learning, it is vital that researchers are aware of their own privilege and how their position could affect their understanding of the intersectional experiences they are investigating.
Intersectionality and its application, irrespective of its context, must be diligently informed by the groups it aims to represent or serve. 
Only when researchers earnestly consider the implications of intersectionality with due gravity and empathy will its application have the sustainable impact we should strive for. 
\subsection{Interdisciplinary Innovation}
The case studies we analyzed all applied intersectionality in machine learning, albeit in different ways. 
We found that some may be using intersectionality to support quantitative approaches while others may be applying it to support the non-technical aspects of developing machine learning technologies.
Despite the differences across the case studies we explored, we found that our guidelines can provide a critical, structured lens for making the necessary considerations for integrating Crenshaw's definition of intersectionality into their machine learning endeavors.
As is known, research is an end-to-end process; from ideation to reporting, we are responsible for the information and insights we distribute through the research community.
We found that many of the case studies in our sample focused on intersectionality in its motivations and approaches, but not nearly as much in their interpretations and discussions. 
This could be, in part, due to the lack of positionality to adequately or accurately provide such insights. 
This emphasizes the importance and value of interdisciplinary collaborations, which studies suggest show the most promise for insight and impact~\cite{blackwell2009radical, villanueva2015collaborative, lepore2023interdisciplinary}.
Consider once more CS3, where the interpretations provided neglected to apply necessary context to properly interpret the findings (or have accurate discussions about what they mean). Collaborations with or insights from those doing work in intersectionality or who have positionality to make these considerations could have provided the necessary context to produce more practical and appropriate insights.
When this is not possible, our work, and resulting guidelines, provide infrastructure and support for holistically integrating intersectionality into the design and development of machine learning technologies and systems.

\subsection{Limitations}

Our efforts aim to form a preliminary foundation for research and practice and to build on when contemplating the application of intersectionality in machine learning pipelines. Hence, the goal of this work is not to provide an exhaustive or comprehensive overview or analysis of intersectionality and ML, but to provide insights on relevant considerations grounded in prior scholarship and example cases. Therefore, we avoid claiming that our guidelines are comprehensive or complete; instead, we focus on their ability to provide guidance that did not previously exist. We anticipate that researchers and practitioners are interested in exploring or utilizing intersectionality for considerations such as fairness or equity in the design and development of machine learning technologies. In our future work, we will conduct a more in-depth analysis to further refine these guidelines. This will support our ability to evaluate and expand  our proposed guidelines with a larger sample of case studies. This includes broadening our scope to include artificial intelligence to create a more robust guide for researchers and practitioners.

\section{Conclusion}
In this paper, we provide a critical analysis of three distinct machine learning projects, each characterized by their application of intersectionality with unique settings and objectives.
To critique these applications, we developed a set of guidelines based on foundational intersectionality scholarship that point to important considerations. 
We found that while existing efforts are generally defining intersectionality accurately and citing relevant scholars, there are important considerations not being made to holistically adopt intersectionality in machine learning methodologies. 
Our goal in conducting this analysis was to enhance the knowledge within the machine learning community regarding the nuanced application of intersectionality and the importance of safeguarding its conceptual integrity across diverse domains.
We encourage future machine learning endeavors delving into intersectionality to prioritize the centrality of power; we provided and discussed concrete guidelines and implications from our efforts that can support the ability to do so in practice. 




\begin{acks}
Thank you Aspiring Scientists Summer Internship Program (ASSIP) program at George Mason University for the opportunity to do this work. 
\end{acks}

\bibliographystyle{ACM-Reference-Format}
\bibliography{references, intersectionality,FAccT}


\end{document}